\documentclass[journal]{IEEEtran}
\IEEEoverridecommandlockouts
\usepackage{cite}
\usepackage{amsmath,amssymb,amsfonts}
\usepackage{algorithmic}
\usepackage{graphicx}
\usepackage{textcomp}
\usepackage{xcolor}
\usepackage{algorithm}
\usepackage{booktabs}
\usepackage{multirow}
\usepackage{makecell}
\usepackage{subfig}
\usepackage{url}
\def\BibTeX{{\rm B\kern-.05em{\sc i\kern-.025em b}\kern-.08em
    T\kern-.1667em\lower.7ex\hbox{E}\kern-.125emX}}

\begin{document}

\title{The Cross-Domain Generalization Cost of Offensive Language Detection}

\author{Ruixing~Ren, Junhui~Zhao,~\IEEEmembership{Senior~Member,~IEEE}, Xiaoke~Sun, Qiuping~Li
	
\thanks{Ruixing Ren, Junhui Zhao are with the School of Electronic and Information Engineering, Beijing Jiaotong University, Beijing 100044, China. (e-mail: renruixing0604@163.com; junhuizhao@hotmail.com)

Xiaoke Sun and Qiuping Li are with the National Computer Network Emergency Response Technical Team/Coordination Center of China (CNCERT/CC), Beijing 100029, China (e-mail: xiaokesun@cert.org.cn; qiupingli\_bj@163.com).
		
	}
	
}


\maketitle

\begin{abstract}
Offensive language detection models generally suffer performance
degradation when deployed across datasets and across languages,
yet most existing studies stop at reporting this phenomenon and
lack a systematic methodology for decomposing the causes of
degradation into attributable components and quantifying the cost
of remediation. This paper proposes a diagnosis and optimization
framework composed of three coordinated technical components.
First, a zero-shot transfer loss decomposition that separates the
performance degradation from OLID to MLMA into two independently
measurable components, namely dataset effect and language effect.
Second, a controlled fine-tuning protocol that quantifies both
adaptation efficiency and the hidden damage inflicted on the source
task by comparing few-shot learning curves under continued
fine-tuning and cold-start starting points. Third, three joint
training strategies incorporating temperature sampling and
experience replay, which offer a controllable Pareto trade-off
between improving multilingual capability and preserving
source-task performance. Experiments built on this framework show
that the dataset effect dominates the zero-shot transfer loss and
substantially outweighs the language effect. Few-shot adaptation
without a replay mechanism, though data-efficient, inflicts source
task damage 4 to 9 times greater than that of the joint training
strategies, and its damage magnitude is highly unstable. The three
joint training strategies trade 3.2 to 4.1 percentage points of
source-task performance for 8.1 to 42.6 percentage points of
multilingual capability gain, forming a clear and controllable
Pareto trade-off. Four methodologically distinct experiments
converge on the same conclusion: the dominant factor governing
cross-lingual generalization is the class-balance of the target
language data itself, rather than the typological distance between
languages. Relevant code can be found in https://github.com/renruixing/The-Cross-Domain-Generalization-Cost-of-Offensive-Language-Detection
\end{abstract}


\section{Introduction}\label{Sec1}

The global spread of social media has made the automatic detection
of offensive language, hate speech, and related harmful content a
core technical component of content safety
governance~\cite{COLD,SWSR}. In practical deployment, however, a
detection model rarely encounters only inputs drawn from the same
distribution and the same language as its training
data~\cite{RenDCAN}. Content moderation systems typically need to
serve multiple language communities simultaneously and to cover
user-generated content from different platforms, whereas
high-quality labeled data available for training tends to be
concentrated in a single dataset and a single language. The
reliability of a model once transferred to a new data source or a
new language directly determines whether the system can be put to
real use.

This deployment gap has attracted some study in the offensive
language detection literature, but existing research falls into
three relatively independent strands that seldom intersect. The
first strand concerns cross-dataset generalization. Swamy et
al.~\cite{swamy2019studying} conducted cross-evaluation across
four English offensive-language datasets and reported macro-F1
drops ranging from 2 to 30 percentage points, spanning a wide
range. Nejadgholi and Kiritchenko~\cite{nejadgholi2020cross}
attribute this drop to two causes: topic bias, referring to
systematic differences in collection channels and topic
distributions across datasets, and task-definition bias, referring
to the fact that the operational definition of offensive content
itself differs across datasets. The systematic review by Yin and
Zubiaga~\cite{yin2021towards} further notes that most work focuses
on reporting the phenomenon, with relatively little examination of
whether the performance gap can be repaired.

The second strand concerns cross-lingual transfer itself.
Multilingual BERT, or mBERT~\cite{devlin2019bert}, is generally
considered to derive its zero-shot cross-lingual transfer ability
from shared subword vocabularies and similar syntactic structures
rather than from explicit cross-lingual alignment
supervision~\cite{pires2019multilingual}.
Nozza~\cite{nozza2021exposing} performed zero-shot hate speech
transfer experiments on English, Italian, and Spanish and found
that models misclassify target-language taboo words as hate
signals, indicating that zero-shot transfer cannot be used
directly and requires deliberate design. Pamungkas et
al.~\cite{pamungkas2021joint} attempted to mitigate this issue via
a joint learning approach with multilingual lexical knowledge
injection, but did not discuss the impact of this method on the
performance of English as the source-language task.

The third strand rarely intersects with the previous two. Conneau
et al.~\cite{conneau2020unsupervised} introduced in their XLM-R
study the notion of the curse of multilinguality, pointing out
that when a model's parameter capacity is fixed, covering more
languages dilutes the effective representational capacity
allocated to any single language; however, this cost has never
been directly measured on concrete tasks such as offensive
language detection. R{\"o}ttger et
al.~\cite{rottger2022data} systematically studied, across five
non-English languages, the amount of target-language fine-tuning
data required to extend offensive language detection to
low-resource languages, finding that a small amount of
target-language data suffices to achieve strong performance and
that gains decay exponentially with additional data; but the study
does not examine the impact of this fine-tuning process on the
performance of English as the source-language task. Work in this
direction continues; Ghorbanpour et
al.~\cite{ghorbanpour2025dataefficient} propose augmenting a small
amount of target-language labeled data via cross-lingual
nearest-neighbor retrieval, likewise without addressing the cost
to source-language performance. Catastrophic forgetting itself was
first systematically formulated by
French~\cite{french1999catastrophic}, and the continual learning
survey by De Lange et al.~\cite{delange2021continual} further
shows that experience-replay based methods can mitigate the damage
inflicted on the original task when a model continues training on
a new distribution; but this mechanism has not previously been
introduced into the cross-lingual training setting for offensive
language detection.

Cross-dataset generalization research asks what performance drops
should be attributed to; cross-lingual transfer and joint training
research asks whether remediation is effective and at what cost;
but none of the three strands provides a framework that examines
both questions together. Cross-dataset generalization research
supplies attribution concepts but lacks a quantitative
decomposition method; cross-lingual transfer research reports
zero-shot failure but does not systematically examine the benefits
and costs of few-shot remediation; the curse of multilinguality
and experience replay both enjoy theoretical support but have
never been directly quantified and compared within the same
experimental design. Table~\ref{tab:comparison} compares
representative prior work with this paper along six dimensions.

\begin{table*}[t]
\centering
\caption{Comparison with representative prior work}
\label{tab:comparison}
\renewcommand{\arraystretch}{1.2}
\begin{tabular}{lcccccc}
\toprule
Work & \makecell{Cross-\\dataset} & \makecell{Cross-\\lingual} & \makecell{Effect\\decomposition} & \makecell{Few-shot\\recoverability} & \makecell{Forgetting\\quantification} & \makecell{Joint training\\comparison} \\
\midrule
Swamy et al.~\cite{swamy2019studying} & \checkmark & $\times$ & $\times$ & $\times$ & $\times$ & $\times$ \\
Nejadgholi \& Kiritchenko~\cite{nejadgholi2020cross} & \checkmark & $\times$ & Conceptual & $\times$ & $\times$ & $\times$ \\
Nozza~\cite{nozza2021exposing} & $\times$ & \checkmark & $\times$ & $\times$ & $\times$ & $\times$ \\
R{\"o}ttger et al.~\cite{rottger2022data} & $\times$ & \checkmark & $\times$ & \makecell{Data-size\\analysis} & $\times$ & $\times$ \\
Pamungkas et al.~\cite{pamungkas2021joint} & $\times$ & \checkmark & $\times$ & $\times$ & $\times$ & \makecell{Knowledge\\injection \checkmark} \\
This work & \checkmark & \checkmark & \checkmark & \checkmark & \checkmark & \makecell{Three\\strategies \checkmark} \\
\bottomrule
\end{tabular}
\end{table*}

Before turning to the proposed framework, it is worth addressing a
question raised by the recent progress of prompting-based large
language models (LLMs): does simply prompting an LLM sidestep the
gap identified above? The evidence suggests otherwise.
Plaza-del-Arco et al.~\cite{plazadelarco2023respectful} showed that
zero-shot prompting with LLMs can achieve performance comparable to
fine-tuned models on some hate speech benchmarks, and subsequent
studies have extended this direction to multilingual and
low-resource settings~\cite{guo2023investigation}. However, the
large-scale evaluation of Edwards and
Camacho-Collados~\cite{edwards2024language} across 16 text
classification datasets found that fine-tuned masked language
models still outperform few-shot prompted generative models on most
tasks, and Fasching and Lelkes~\cite{fasching2025model} documented
substantial inconsistencies in moderation decisions across
LLM-based systems, neither of which offers the dataset-versus-language
decomposition or the adaptation-cost quantification this paper
targets. LLM-based moderation also entails substantially higher
inference cost per input, a significant obstacle for pipelines that
must score every incoming post. For these reasons, compact
fine-tuned encoders remain the deployment-relevant regime, and the
framework below is developed within that regime.

To address this gap, this paper proposes an integrated framework
of diagnosis, adaptation, and optimization, taking as its
foundation a three-level cascaded detection model trained on the
English offensive language dataset
OLID~\cite{zampieri2019olid}, and systematically characterizes
and mitigates its cross-dataset and cross-lingual generalization
loss on the multilingual hate speech dataset
MLMA~\cite{ousidhoum2019mlma}, which covers English, French, and
Arabic. The main contributions of this paper are summarized as
follows.

\begin{itemize}
  \item A diagnostic experimental design that, by controlling for
    the language variable, decomposes the zero-shot transfer
    performance loss into a dataset-effect component and a
    language-effect component, rather than merely reporting the
    aggregated drop in performance.
  \item By contrasting few-shot learning curves from a
    continued-fine-tuning start on the trained model with those
    from a cold-start on the raw pretrained model, this paper
    reveals that few-shot adaptation, while improving
    target-domain performance, inflicts hidden damage on
    source-task performance that is far greater than that caused
    by joint-training methods, and that this damage is highly
    unstable.
  \item By comparing three language-joint-training strategies,
    this paper for the first time quantifies through direct
    controlled experiments the specific cost of the curse of
    multilinguality on the offensive language detection task, and
    provides a basis for strategy selection oriented toward
    real-world deployment.
\end{itemize}

The remainder of this paper is organized as follows.
Section~\ref{Sec2} describes the datasets, the baseline model, and
the design of the three experiments. Section~\ref{Sec3} presents
the experimental results and analysis. Section~\ref{Sec4}
concludes the paper.

\section{Method}\label{Sec2}

\subsection{Method Overview}\label{Sec2.0}

The experimental design of this paper is organized around a common
starting point: a baseline cascaded model trained on the English
OLID data, whose training details are given in
Section~\ref{Sec2.2}, followed by an examination of the model's
behavior on the multilingual MLMA data from three complementary
angles. The first angle is the diagnosis of
Section~\ref{Sec2.3}, which uses only the zero-shot evaluation
results of the baseline model to decompose the performance loss
into a dataset-effect component and a language-effect component,
answering the question of where the loss comes from. The second
angle is the adaptation of Section~\ref{Sec2.4}, which builds on
the diagnostic results to ask whether the loss can be repaired
with a small amount of target-language data and at what cost, by
contrasting the learning curves under a continued-fine-tuning
start and a cold-start. The third angle is the optimization of
Section~\ref{Sec2.5}, which steps outside the few-shot setting and
trains three joint strategies using the full candidate pool,
asking how, if the goal is not merely to patch the loss but to
positively acquire a usable multilingual capability, one should
train, and at what cost. The three angles share the same data
partition described in Section~\ref{Sec2.1} and the same
evaluation metric described in Section~\ref{Sec2.6}, so that the
results of the diagnosis, adaptation, and optimization experiments
can be placed in the same table for direct comparison, which is
also the prerequisite for the cross-verification of
Section~\ref{Sec3.5}.

\subsection{Datasets and Label Mapping}\label{Sec2.1}

This paper uses two real, publicly available datasets.
\textbf{OLID}~\cite{zampieri2019olid} is an English offensive
language dataset containing 13{,}203 examples after cleaning,
annotated under a three-level scheme: subtask A determines whether
the content is offensive, with labels OFF or NOT; subtask B
determines whether the offense is directed at an identifiable
target, with labels TIN or UNT; subtask C determines the type of
target, with labels IND, GRP, or OTH, corresponding to individual,
group, and other. This paper splits the training and validation
sets by 9:1 stratified sampling; the official test set is used for
final reporting, containing 860 examples for subtask A and 213 for
subtask C.

\textbf{MLMA}~\cite{ousidhoum2019mlma} is a hate speech dataset
covering English, French, and Arabic, with 5{,}586, 4{,}006, and
3{,}350 examples respectively after cleaning. The original
annotation fields include the hostility-type field
\texttt{sentiment}, whose value is a multi-label combination drawn
from normal, offensive, abusive, hateful, disrespectful, and
fearful, and the target-group field \texttt{group}, whose values
include individual, other, and specific group names. Because
MLMA's annotation scheme does not naturally align with OLID's
three-level scheme, this paper makes only the following mapping:

\begin{itemize}
  \item Subtask A: the \texttt{sentiment} field is split into a
    label set; if the set equals the singleton normal it is mapped
    to NOT, otherwise it is mapped to OFF.
  \item Subtask C: applied only to examples mapped to OFF; the
    \texttt{group} field value individual is mapped to IND, other
    is mapped to OTH, and all remaining specific group names are
    uniformly mapped to GRP.
  \item Subtask B is not mapped: MLMA was collected via
    identity-attribute keyword retrieval, so its examples
    naturally carry an identifiable target and lack genuine
    instances corresponding to non-targeted generic profanity;
    forcing a mapping would introduce label noise, and therefore
    all experiments in this paper concern only subtasks A and C.
\end{itemize}

\begin{figure}[t]
\centering
\includegraphics[width=0.95\linewidth]{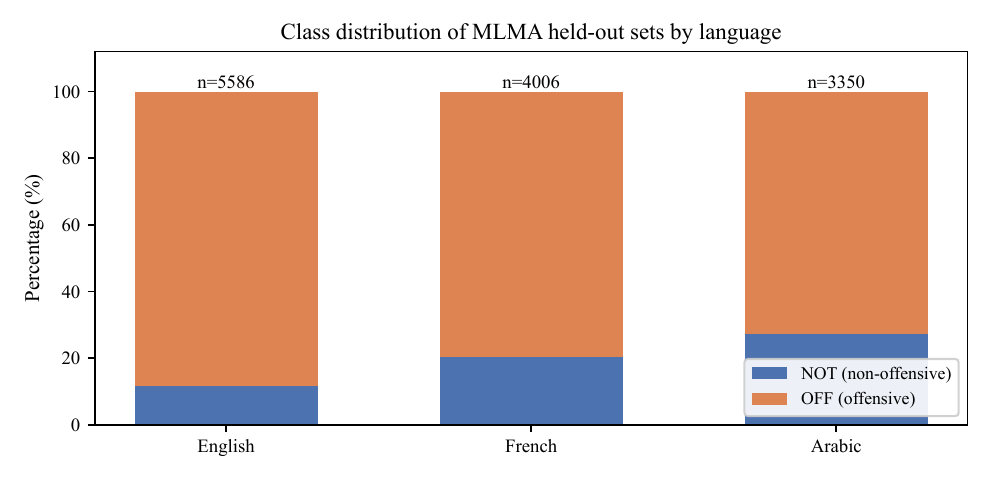}
\caption{Class distributions of the three-language MLMA held-out
sets. The class imbalance decreases from English to French to
Arabic.}
\label{fig:mlma_dist}
\end{figure}

To ensure that the results of the diagnostic evaluation, the
few-shot adaptation, and the joint training in this paper can be
directly compared, this paper applies a uniform partition to each
language of MLMA: with a fixed random seed and stratified sampling
by subtask A, the data of each language is split in half, with one
half designated as the held-out set that never participates in
training in any experiment and on which all conditions are
finally evaluated, and the other half designated as the candidate
pool. Fig.~\ref{fig:mlma_dist} shows the class distributions of
the three-language held-out sets, with the sample size $n$
denoting the total number of examples for each language. The NOT
proportions of the three-language held-out sets are 11.8\%,
20.4\%, and 27.3\% respectively; the ordering of this imbalance is
repeatedly confirmed in Section~\ref{Sec3} across multiple
independent experiments as the decisive variable governing
cross-lingual generalization.

\subsection{Backbone and Baseline Cascaded Model}\label{Sec2.2}

This paper adopts \texttt{bert-base-multilingual-cased}, i.e.\
mBERT, as the backbone model. All three research questions in this
paper require the model to be simultaneously capable of
understanding English and understanding French and Arabic; the
former is used to inherit the offensive-language discrimination
learned from OLID, and the latter is used to evaluate and train
cross-lingual generalization. A monolingual BERT does not have the
vocabulary coverage to process non-English text and therefore does
not satisfy this prerequisite. The cross-lingual ability of mBERT
is obtained at the cost of some monolingual task performance; the
specific magnitude of this cost is quantified in the experiments
of Section~\ref{Sec3.4}.

\begin{figure}[t]
\centering
\includegraphics[width=\linewidth]{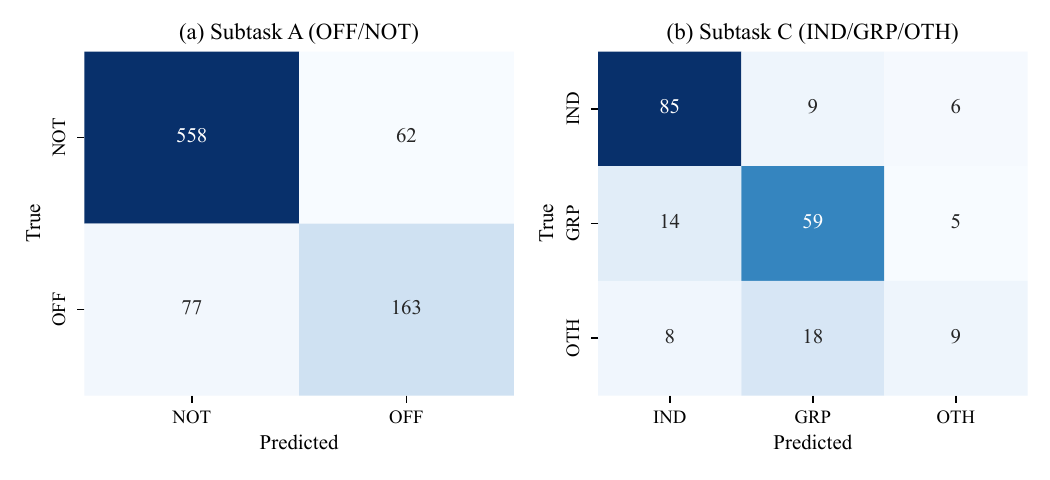}
\caption{Confusion matrix of the baseline cascaded model on the
OLID official test set.}
\label{fig:baseline_cm}
\end{figure}

The baseline model is a two-stage cascaded classifier: the subtask
A head determines OFF or NOT over all training examples; the
subtask C head is trained only on examples whose subtask A label
is OFF and whose subtask B label is TIN, and determines IND, GRP,
or OTH. The two classification heads share the \texttt{[CLS]}
representation $h \in \mathbb{R}^{768}$ produced by the mBERT
encoder, and each is followed by a fully connected layer that
maps $h$ to a class probability distribution for the corresponding
subtask:
\begin{equation}
z = W h + b, \qquad p_i = \frac{e^{z_i}}{\sum_{j=1}^{K} e^{z_j}}
\label{eq:softmax}
\end{equation}
where $K$ is the number of classes for that subtask, taken as 2
for subtask A and 3 for subtask C. This probability distribution
is then used to compute the training loss and drives the parameter
update. Training follows the common fine-tuning practice for
BERT-style models on text classification
tasks~\cite{sun2019finetune}: the AdamW~\cite{loshchilov2019adamw}
optimizer is used together with discriminative learning
rates~\cite{howard2018ulmfit}, with $2\times10^{-5}$ for the
encoder and $1\times10^{-4}$ for the classification heads, coupled
with a linear warm-up over the first 10\% of training steps
followed by linear decay. Training is length-controlled by early
stopping with validation Macro-F1 as the monitored metric and a
patience of 3. Dropout~\cite{srivastava2014dropout} is applied
before each classification head to mitigate overfitting. Given
that the class distributions of subtasks A and C are unbalanced to
different degrees (subtask A about 2:1, and OTH being the minority
class in subtask C), $\mathbf{p}$ is fed into a class-weighted
cross-entropy loss:
\begin{equation}
\mathcal{L} = -\sum_{c=1}^{K} w_c\, y_c \log p_c, \qquad
w_c = \frac{N}{K \times n_c}
\label{eq:weighted_ce}
\end{equation}
where $N$ is the total number of training examples and $n_c$ is
the number of examples in class $c$. After training, the baseline
model achieves Macro-F1 of 0.795 on subtask A and 0.623 on subtask
C on the OLID official test set; its confusion matrix is shown in
Fig.~\ref{fig:baseline_cm}. Its weights are then kept fixed and
serve as the common zero-shot starting point for all experiments
in Sections~\ref{Sec2.3} and~\ref{Sec2.4}.

\subsection{Diagnostic Experiment: Separating Dataset Effect from
Language Effect}\label{Sec2.3}

With the weights of the baseline model trained in
Section~\ref{Sec2.2} kept fixed, this section evaluates the model
in a zero-shot manner directly on the three-language MLMA held-out
sets, without any additional training, in order to answer the
question of where the performance loss comes from. Let
$S(\cdot)$ denote the model's Macro-F1 on a given test set, let
$D_{\text{OLID}}$ denote the OLID official test set, and let
$D_{\text{MLMA}}^{L}$ denote the MLMA held-out set of language
$L$. Define the total performance drop as
\begin{equation}
\Delta_{\text{total}}(L) = S(D_{\text{OLID}}) - S(D_{\text{MLMA}}^{L})
\label{eq:delta_total}
\end{equation}
For English, the language is the same as the training language and
only the dataset differs, so
$\Delta_{\text{total}}(\text{en})$ is fully attributable to the
dataset effect; denote
$\Delta_{\text{dataset}} = \Delta_{\text{total}}(\text{en})$. For
French and Arabic, an additional cross-lingual transfer cost is
superimposed on top of the dataset effect:
\begin{equation}
\Delta_{\text{language}}(L) = \Delta_{\text{total}}(L) - \Delta_{\text{dataset}}, \quad L \in \{\text{fr}, \text{ar}\}
\label{eq:delta_lang}
\end{equation}
If $\Delta_{\text{dataset}}$ accounts for the majority of
$\Delta_{\text{total}}(L)$, then differences between datasets and
annotation standards are the dominant factor; the magnitude of
$\Delta_{\text{language}}(L)$ reflects the true cross-lingual
transfer cost. This decomposition can be computed from the
zero-shot evaluation of the baseline model alone and requires no
additional training.

\subsection{Few-Shot Adaptation Experiment}\label{Sec2.4}

The decomposition in Section~\ref{Sec2.3} answers the question of
where the loss comes from; this section further asks whether the
loss can be repaired. To test the recoverability of the
performance gap, this paper contrasts two conditions along the
learning curves under different adaptation-data sizes $N$. This
design is aligned with the perspective of R{\"o}ttger et
al.~\cite{rottger2022data}, who study the gains from
target-language fine-tuning-data volume; the difference is that
this paper also tracks the change in performance on English as
the source-language task, rather than reporting only
target-language performance. Under the continued-fine-tuning
condition, the subtask A head of the baseline cascaded model
serves as the starting point, and only $N$ MLMA examples sampled
from the candidate pool are used to continue fine-tuning, with
evaluation on the held-out set. Under the cold-start condition,
the raw pretrained mBERT is used as the starting point, and only
the same $N$ examples are used for fine-tuning, with evaluation
on the same held-out set. $N$ takes values 0, 50, 200, and 500,
where $N{=}0$ is the zero-shot evaluation and mutually corroborates
the diagnostic experiment of Section~\ref{Sec2.3}. Each combination
of language and $N$ is independently repeated 3 times, with
re-stratified sampling and a fresh random seed, and the mean and
standard deviation are reported. If the continued-fine-tuning
condition significantly outperforms the cold-start condition at
small sample sizes, it indicates that the discriminative ability
learned by the baseline model is transferable and that the
performance gap arises mainly from differences in annotation
standards rather than from insufficient model capacity.

\subsection{Joint Training Strategies}\label{Sec2.5}

The few-shot adaptation in Section~\ref{Sec2.4} answers whether
the performance gap can be repaired with a small amount of data,
but the few-shot setting itself sacrifices most of the data in the
candidate pool. This section moves beyond that limitation; all
three strategies take the entire candidate pool, rather than a
restricted few-shot subset, as available training data, with the
goal of obtaining a unified model whose source-task performance
does not degrade significantly and that simultaneously exhibits a
practically usable multilingual capability.

The first strategy is direct concatenation: the OLID training set
and the MLMA candidate pools of the three languages are merged
directly, shuffled, and then used for training. English, together
with OLID, accounts for about 81\% of the merged training data,
and the presence of French and Arabic is significantly diluted.

The second strategy is language-balanced sampling. The data
composition is the same as in direct concatenation, but during
training, examples are grouped by language and the probability
with which each language appears in a batch is controlled by
temperature sampling~\cite{conneau2020unsupervised}:
\begin{equation}
q_L = \frac{n_L^{1/T}}{\sum_{L'} n_{L'}^{1/T}}
\label{eq:temp_sampling}
\end{equation}
where $n_L$ is the number of examples in language $L$ and $T$ is
the temperature parameter, set to $T{=}5$ in this paper,
consistent with the XLM-R paper. The actual sampling weight of
each example is $q_L / n_L$, and batch-level language reweighting
is implemented via a weighted random sampler.

The third strategy is two-stage training with experience
replay~\cite{rolnick2019experience}. The first stage reuses the
baseline model without retraining; the second stage continues
training solely on the full MLMA candidate pool, and, to mitigate
the catastrophic forgetting observed in Section~\ref{Sec2.4}, an
additional proportion $r{=}0.3$ of OLID training examples is mixed
in as experience replay:
\begin{equation}
\mathcal{D}_{\text{stage2}} = \mathcal{D}_{\text{MLMA}} \cup \mathcal{D}_{\text{replay}}, \quad
|\mathcal{D}_{\text{replay}}| = r \cdot |\mathcal{D}_{\text{MLMA}}|
\label{eq:replay}
\end{equation}
All three strategies are evaluated on the OLID official test set
and on the three-language MLMA held-out sets, and their results
are placed in the same table as the results of the diagnostic
experiment and the few-shot adaptation experiment for comparison.

\subsection{Evaluation Metric}\label{Sec2.6}

Because the class distributions of all three test sets are
unbalanced, plain accuracy is prone to being dominated by the
majority class and losing discriminative power; this paper
therefore uses macro-averaged F1, i.e.\ Macro-F1, as the core
metric for both subtasks A and C:
\begin{equation}
\text{F1}_{\text{macro}} = \frac{1}{K}\sum_{i=1}^{K}
\frac{2\,\text{Precision}_i \cdot \text{Recall}_i}{\text{Precision}_i + \text{Recall}_i}
\label{eq:macrof1}
\end{equation}

\section{Experiments and Results}\label{Sec3}

\subsection{Experimental Setup}\label{Sec3.1}

\begin{figure}[t]
\centering
\includegraphics[width=\linewidth]{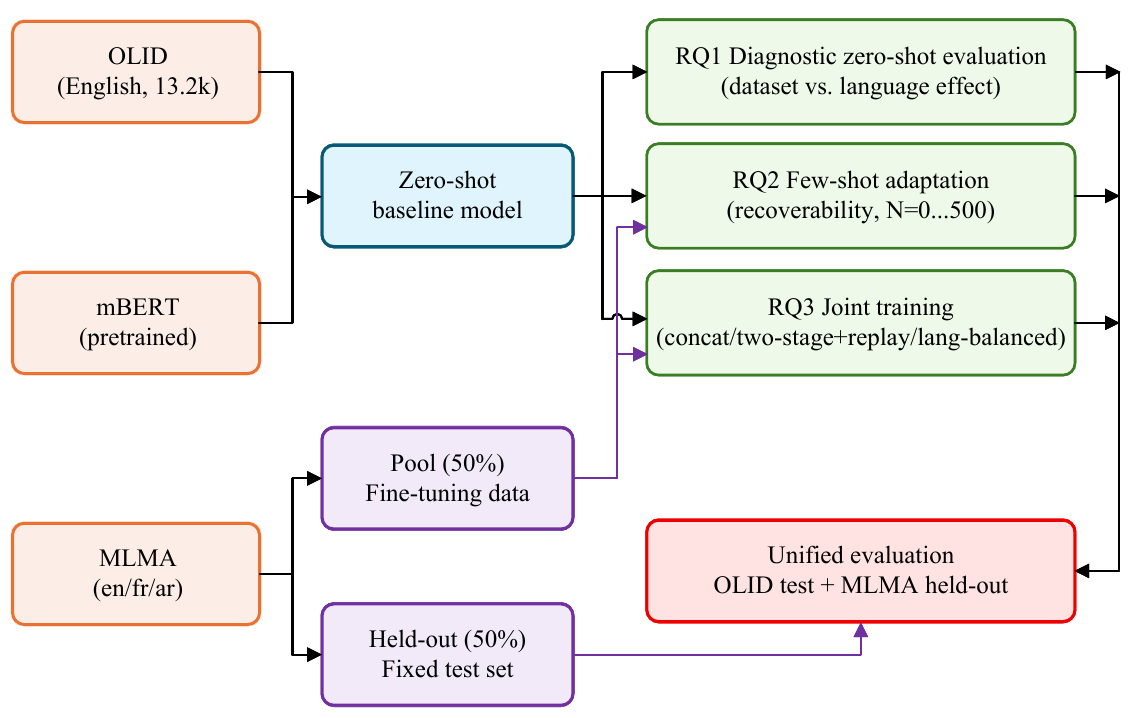}
\caption{The overall experimental framework of this paper. The
diagnostic evaluation, few-shot adaptation, and joint training
experiments share the same baseline model and the same fixed
held-out partition. The candidate pool is used only as
fine-tuning data for the few-shot adaptation and joint training
experiments and does not participate in the diagnostic
evaluation.}
\label{fig:framework}
\end{figure}

Experiments are carried out on a single NVIDIA RTX 4060 Ti with
16GB of memory, running PyTorch 2.3.0+cu121. The hyperparameters
shared by the three joint-training strategies are: batch size 16,
maximum sequence length 128, maximum training epochs 10, early
stopping patience 3, and FP16 mixed precision enabled throughout
training. Fig.~\ref{fig:framework} gives the overall experimental
framework of this paper: after the baseline model is trained, it
is used for three groups of experiments, namely diagnostic
evaluation, few-shot adaptation, and joint training. The three
groups share the same fixed MLMA held-out sets to ensure strictly
comparable results.

\begin{figure*}[t]
	\centering
	\includegraphics[width=\linewidth]{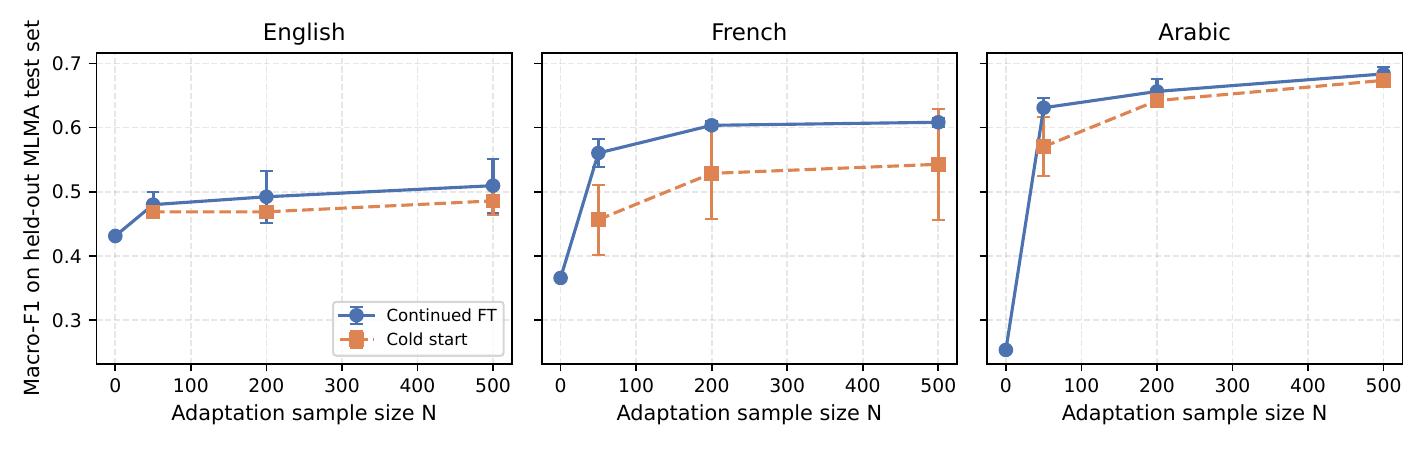}
	\caption{Learning curves of the few-shot adaptation experiment;
		error bars are the standard deviation over 3 repetitions. The
		continued-fine-tuning condition consistently outperforms the
		cold-start condition; the English side repeatedly falls into
		degenerate solutions with zero standard deviation at $N{\le}200$.}
	\label{fig:adaptation}
\end{figure*}

\subsection{Results of the Diagnostic Experiment}\label{Sec3.2}

The first row of Table~\ref{tab:main_results} reports the
zero-shot evaluation results of the baseline model. The English
held-out set uses the same language as the training data, differing
only in the source dataset; its Macro-F1 drops from 0.795 on the
OLID test set to 0.431, giving
$\Delta_{\text{dataset}}=0.364$. On top of this, the scores on
French and Arabic drop further to 0.366 and 0.254 respectively,
giving $\Delta_{\text{language}}(\text{fr})=0.065$ and
$\Delta_{\text{language}}(\text{ar})=0.177$. The dataset effect of
0.364 far exceeds the language effect of 0.065 to 0.177,
indicating that the performance loss of zero-shot transfer arises
mainly from systematic differences in annotation standards
between OLID and MLMA rather than from the transfer cost of the
language itself. This is mechanistically consistent with the
task-definition bias discussed in Section~\ref{Sec1}: MLMA is
collected via identity-attribute keyword retrieval, and its
operational definition of offensiveness covers a large amount of
indirect discriminatory expression without explicit profanity,
which differs systematically from OLID's annotation orientation
that emphasizes direct profanity and insults.

\subsection{Results of the Few-Shot Adaptation Experiment}\label{Sec3.3}

Fig.~\ref{fig:adaptation} shows the learning curves for the three
languages. Overall, the continued-fine-tuning condition is
consistently better than the cold-start condition at small
sample sizes, indicating that the discriminative ability learned
by the baseline model is indeed transferable. The learning curves
on the English side are markedly less stable than those on
French and Arabic: at $N{=}50$ and $N{=}200$, the cold-start
condition repeatedly degenerates into a degenerate solution that
predicts the majority class, with a standard deviation of 0 and
three repetitions yielding identical predictions; the
continued-fine-tuning condition also partially exhibits the same
degeneracy, and only at $N{=}500$ do the learning curves of all
three languages stably escape the degenerate solution. This
phenomenon directly corresponds to the class imbalance of the
English held-out set, whose NOT-to-OFF ratio is 11.8\% to 88.2\%,
the most imbalanced among the three languages, indicating that the
training stability of few-shot adaptation is likewise governed by
the class balance of the target-domain data.

To probe the cost of this adaptation on the source task, we further
measured, at $N{=}500$, the impact of the continued-fine-tuning
condition on the OLID official test set; see the second row of
Table~\ref{tab:main_results}. Adaptation on the three languages
reduces the OLID test-set Macro-F1 to $0.435\pm0.226$,
$0.530\pm0.053$, and $0.655\pm0.041$ respectively, with drops of 14
to 36 percentage points, far exceeding the 3.2 to 4.1
percentage-point drops of the three joint-training strategies. The
standard deviation for English reaches 0.226, meaning that the three
specific repetitions, 0.371, 0.686, and 0.248, fluctuate by more
than a factor of three. The fundamental cause of this difference is
that the fine-tuning stage of few-shot adaptation contains no
original OLID data at all, whereas the three joint training
strategies each retain a substantial proportion of the original
distribution; this is discussed in detail in the next subsection.

\subsection{Comparison of Joint Training Strategies}\label{Sec3.4}

\begin{figure}[t]
\centering
\includegraphics[width=\linewidth]{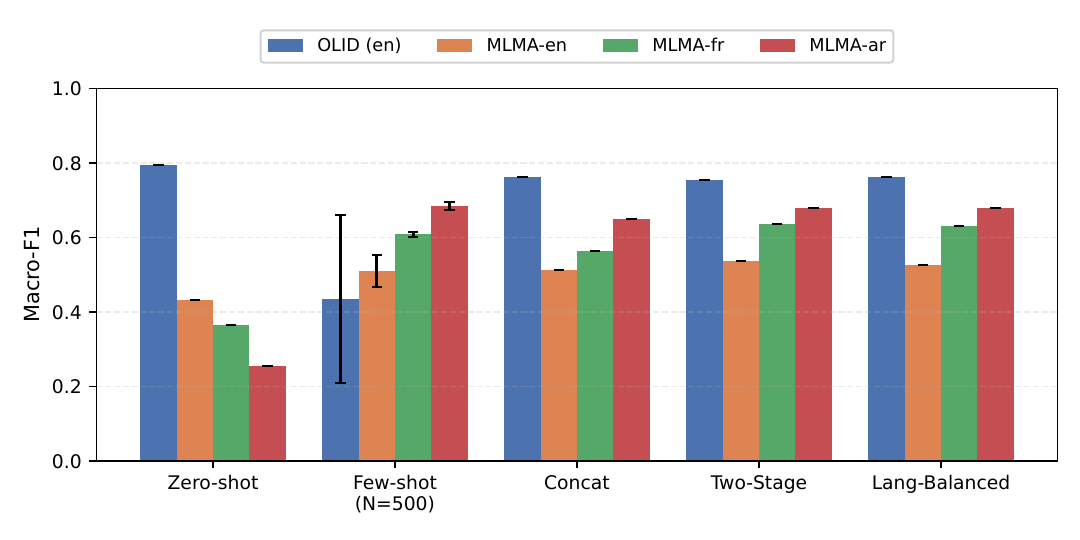}
\caption{Macro-F1 comparison of the five conditions on the four
test sets. Error bars are the standard deviation over 3
repetitions for the few-shot adaptation condition; the other
conditions correspond to single runs.}
\label{fig:main_results}
\end{figure}

Table~\ref{tab:main_results} and Fig.~\ref{fig:main_results}
present the same set of results. The three joint training
strategies all incur drops of 3.2 to 4.1 percentage points on
OLID, providing the first quantification of the cost of the curse
of multilinguality on this task. The two-stage replay strategy
suffers the largest drop, whose OFF recall correspondingly falls
from 67.9\% to 50.0\%, indicating that the decision boundary
shifts toward predicting NOT. The gains on the three languages
show a consistent ordering: Arabic is the largest, at 0.396 to
0.426; French is intermediate, at 0.197 to 0.271; and English is
the smallest, at 0.081 to 0.107. This ordering corresponds exactly
to the NOT proportions in the held-out sets, 11.8\%, 20.4\%, and
27.3\%, and mutually corroborates the results of
Sections~\ref{Sec3.2} and~\ref{Sec3.3}, converging on the same
conclusion: the class balance of the target-language data itself
is the dominant factor governing cross-lingual generalization.
Direct concatenation and two-stage replay form a Pareto trade-off:
the latter pays about 0.9 percentage points more in English cost
in exchange for an additional average gain of about 4.3
percentage points across the three languages, which can be
selected according to the priorities of English versus
multilingual capability in the deployment scenario.

\begin{figure}[t]
\centering
\includegraphics[width=\linewidth]{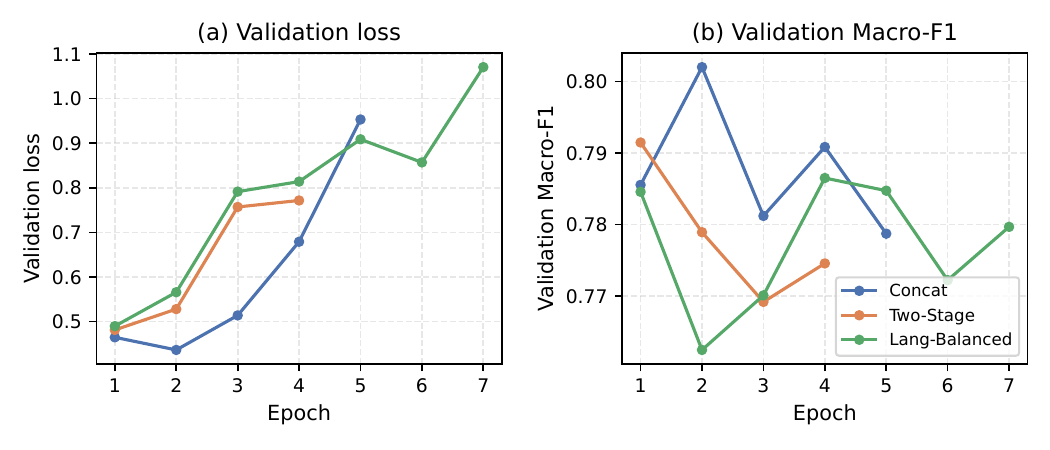}
\caption{Validation loss (subplot a) and validation Macro-F1
(subplot b) of the three joint training strategies as functions
of training epoch.}
\label{fig:joint_curves}
\end{figure}

Fig.~\ref{fig:joint_curves} shows the training curves of each of
the three joint training strategies. For direct concatenation, the
validation loss reverses from decreasing to increasing after
epoch 2, while the training loss keeps decreasing, producing a
clear scissors gap between the two curves that is a hallmark of
overfitting. For two-stage replay, the starting point is already a
well-converged baseline model, so the marginal returns of
continued training are rapidly exhausted, resulting in the
smallest number of early-stopping epochs, only 4. Language-balanced
sampling, because of the higher single-example repetition rate
brought about by the weighted random sampler, has the greatest
fluctuation during training, with validation loss oscillating
between 0.490 and 1.070; its early-stopping epoch count is the
largest, reaching 7, making it the least stable of the three
strategies during training, an engineering cost that must be
additionally weighed when this strategy is chosen.

\begin{table*}[t]
\centering
\caption{Main experimental results. Values are Macro-F1; for
non-zero-shot conditions, the second row of each block reports the
delta from the zero-shot baseline. All few-shot adaptation and joint
training results are evaluated on the same fixed held-out set.}
\label{tab:main_results}
\renewcommand{\arraystretch}{1.2}
\begin{tabular}{lcccc}
\toprule
Condition & OLID English & MLMA English & MLMA French & MLMA Arabic \\
\midrule
Zero-shot & 0.795 & 0.431 & 0.366 & 0.254 \\
Few-shot $N{=}500$ & 0.435 & 0.510 & 0.608 & 0.684 \\
 & $-$0.360 & $+$0.079 & $+$0.242 & $+$0.430 \\
Direct concatenation & 0.763 & 0.512 & 0.563 & 0.650 \\
 & $-$0.032 & $+$0.081 & $+$0.197 & $+$0.396 \\
Two-stage replay & 0.754 & 0.538 & 0.637 & 0.679 \\
 & $-$0.041 & $+$0.107 & $+$0.271 & $+$0.425 \\
Language-balanced & 0.763 & 0.526 & 0.630 & 0.680 \\
 & $-$0.032 & $+$0.095 & $+$0.264 & $+$0.426 \\
\bottomrule
\end{tabular}
\end{table*}

\subsection{Cost Comparison of Few-Shot Adaptation and Joint
Training}\label{Sec3.5}

When the results of Sections~\ref{Sec3.3} and~\ref{Sec3.4} are
compared along the dimension of whether the source task is
harmed, a conclusion emerges that was not anticipated at the
start of the experimental design: few-shot adaptation requires
far less data and training cost than joint training, on the order
of 500 examples versus thousands to more than ten thousand, yet
its damage to the source task is 4 to 9 times that of the joint
training strategies, and this damage magnitude is itself highly
unstable. The fundamental cause of this difference is that the
fine-tuning stage of few-shot adaptation mixes in no original
training data at all, whereas each of the three joint training
strategies retains a substantial proportion of the original
distribution: English accounts for 81\% in direct concatenation,
two-stage replay explicitly sets 30\% of OLID rehearsal samples,
and after temperature adjustment the language-balanced sampling
still targets 43\% for English. This is consistent with existing
understanding in the continual learning literature on the
effectiveness of experience
replay~\cite{delange2021continual}: regardless of the specific
data organization strategy, retaining a certain proportion of
the original training data during training is a necessary
condition for mitigating catastrophic forgetting.

\subsection{Limitations}\label{Sec3.6}

Several limitations of this study should be acknowledged. First, the
cross-lingual analysis covers three languages, English, French, and
Arabic, from two language families; while these three languages span
a range of class-imbalance conditions that is sufficient to establish
the ordering effect reported in this paper, extending the analysis to
a broader typological sample would strengthen the generality of the
class-balance conclusion. Second, the label mapping from MLMA to the OLID three-level
scheme is an approximation and does not cover subtask B, so the
cross-dataset analysis is confined to subtasks A and C. Third, all
experiments in this paper use the mBERT backbone; whether the same
decomposition and cost trade-off carry over to more recent
multilingual encoders such as XLM-R~\cite{conneau2020unsupervised} or
to LLM-based approaches is an open question left for future
investigation.

\section{Conclusion}\label{Sec4}

Starting from an offensive language detection cascaded model
trained on OLID, this paper systematically examined its
cross-dataset and cross-lingual generalization through three
progressive empirical studies: a diagnostic experiment, a few-shot
adaptation experiment, and a comparison of joint training
strategies. The results showed that the performance drop under
zero-shot transfer originated primarily from differences in
datasets and annotation standards rather than from the language
itself; few-shot adaptation without an experience-replay
mechanism, although data-efficient, inflicted source-task damage
that was far greater than that of joint training and highly
unstable; and each of the three joint training strategies traded a
quantifiable cost in source-task performance for gains in
multilingual capability, forming a clear Pareto trade-off. Four
methodologically distinct experiments converged on the same
conclusion: the dominant factor governing cross-lingual
generalization was the class balance of the target-language data
itself rather than the typological distance between languages.
This conclusion provided an actionable engineering basis for
cross-dataset and cross-lingual deployment of offensive language
detection systems: whether new labeled data needed to be collected,
how much data were needed, and whether joint training was worth its
cost can all be judged with reference to the quantitative results
of this paper. Beyond the specific findings, the diagnosis-adaptation-optimization
framework introduced here is not tied to the offensive
language detection task and can in principle be applied to other
cross-dataset, cross-lingual generalization problems where
attributing the loss and quantifying the remediation cost matter,
which is a promising direction for future work.

\bibliographystyle{IEEEtran}

\end{document}